\documentclass[sigconf,noacm]{acmart} % <--- Note the spelling: 'noacm'

\AtBeginDocument{%
  }

\setcopyright{none}
\renewcommand\footnotetextcopyrightpermission[1]{}

\acmConference[ACM CAIS 2026: RLEval Workshop]{ACM CAIS 2026: RLEval Workshop: Methods and Reinforcement Learning Environments for Evaluating AI Agents}{May 26, 2026}{San Jose, CA}

\begin{document}

\title{Beyond Leaderboards: Tokenomics of Agentic Small Language Model Ensembles}

\author{Alexei N. Skurikhin}
\email{alexei@lanl.gov}
\orcid{0000-0001-5606-4933}
\affiliation{%
  \institution{Los Alamos National Laboratory}
  \city{Los Alamos}
  \state{New Mexico}
  \country{USA}
}

\author{Emily M. Taylor}
\email{ecasleton@lanl.gov}
\affiliation{%
  \institution{Los Alamos National Laboratory}
  \city{Los Alamos}
  \state{New Mexico}
  \country{USA}
}

\author{Nathan A. DeBardeleben}
\email{ndebard@lanl.gov}
\affiliation{%
  \institution{Los Alamos National Laboratory}
  \city{Los Alamos}
  \state{New Mexico}
  \country{USA}
}

\renewcommand{\shortauthors}{Skurikhin et al.}

\begin{abstract}
As large language models (LLMs) move from standalone assistants into agentic workflows, evaluation must extend beyond scalar leaderboard accuracy to account for operational reliability, cost, latency, and token efficiency. We use an agentic ensemble of small language models (SLMs) with an SLM-judge-mediated feedback loop as a case study for such beyond-leaderboard evaluation. On the 541-prompt IFEval benchmark, the best ensemble achieves $97.34\%$ strict prompt accuracy, exceeding the strongest standalone LLM baseline, gpt-5.4, by $5.81$ percentage points while operating in a lower-cost regime. We then analyze the tokenomics and operational behavior behind this gain, including cost per sample, token composition, useful-output goodput, feedback-loop recovery, latency decomposition, and performance across instruction categories and constraint counts. Our results show that agentic SLM ensembles can trade additional test-time tokens and orchestration overhead for improved instruction-following fidelity, motivating multi-dimensional evaluation protocols for future agentic AI systems.
\end{abstract}





\maketitle

\section{Introduction}
Large language models (LLMs) are increasingly used as components of agentic workflows, where success depends not only on producing plausible answers but also on reliably satisfying explicit constraints, schemas, and protocol requirements~\cite{liu2023agentbench,alzahrani2024benchmarks,miller2025evaluating,mialon2023gaia,jimenez2023swebench,koh2024visualwebarena}. This shift exposes a limitation of conventional leaderboard-style evaluation: aggregate accuracy alone does not capture whether a system is practical to deploy when iterative feedback, orchestration, latency, and token cost are part of the inference procedure~\cite{sheikhi2026beyond,liang2023helm,chiang2024chatbotarena,chen2023frugalgpt}. Instruction-following benchmarks such as IFEval~\cite{zhou2023ifeval} provide a useful testbed for this setting because they decompose prompts into verifiable constraints involving formatting, length, inclusion/exclusion rules, and multi-instruction composition. At the same time, standalone frontier models can be expensive for repeated agentic inference, while compute-efficient small language models (SLMs) often struggle with complex constraint combinations~\cite{belcak2025slmagents,subramanian2025slmsurvey}.

We therefore evaluate an agentic ensemble of SLMs implemented using Model Context Protocol (MCP) as the integration framework ~\cite{anthropic2024mcp,modelcontextprotocol2025spec}, combining programmatic IFEval checking with an SLM judge-mediated feedback loop. The system first demonstrates that SLM ensembles can be competitive with, and in our experiments outperform, standalone LLM counterparts on instruction-following accuracy. We then shift the analysis beyond leaderboard accuracy to examine the tokenomics and operational behavior that enable this improvement, including cost per sample, relative token composition, useful output goodput, iterative recovery, latency decomposition, category-level performance, and robustness as the number of constraints increases. Our novelty is a multi-dimensional evaluation of agentic SLM ensembles that treats instruction fidelity, token overhead, feedback-loop cost, and deployment efficiency as coupled evaluation targets. The contribution is an empirical case study showing both that SLM ensembles can work well on IFEval and that their benefits cannot be understood from accuracy alone.

\section{Ensemble System Design and Evaluation}

We developed an agentic SLM ensemble implemented using MCP as the communication layer. The system coordinates multiple compute-efficient SLMs as response generators and uses an SLM judge to support tie-breaking and feedback generation. MCP is used to structure the interaction among generators, judge, and the programmatic IFEval checker, enabling reproducible multi-model execution and iterative correction.

For each IFEval prompt, the ensemble runs for at most three attempts. In the first attempt, all SLM ensemble members independently generate candidate responses to the same multi-constraint prompt. Each candidate is then evaluated using the IFEval programmatic checker. If multiple candidates pass, the SLM judge acts as a stylistic tie-breaker among already compliant outputs. If no candidate satisfies all constraints, the judge converts the violations identified by the programmatic checker into targeted, natural-language corrective feedback for the ensemble members. This process is repeated for up to two corrective attempts.

This design separates three functions that are often conflated in aggregate LLM evaluation: generation, constraint verification, and feedback-guided correction. It also supports a principled Best-of-N interpretation for IFEval: because the benchmark evaluates instruction following rather than domain knowledge, selecting the candidate with maximal constraint satisfaction is an appropriate way to measure the best instruction-compliant output available from the ensemble. We benchmark five ensemble configurations against standalone large and small models from OpenAI, Google, Anthropic, Meta, and Mistral. Full model and ensemble compositions are provided in Table~\ref{tab:ensemble_configs}.

\subsection{Beyond-Leaderboard Evaluation}

We evaluate the proposed system on the IFEval benchmark~\cite{zhou2023ifeval}, which contains 541 prompts with rule-based instruction checks. To maintain compatibility with the standard benchmark, we report the primary IFEval accuracy metrics:
\begin{itemize}
    \item \texttt{Strict Prompt-level Accuracy}: the fraction of prompts for which every underlying constraint is satisfied.
    \item \texttt{Strict Instruction-level Accuracy}: the fraction of individual instructions satisfied across the full benchmark.
    \item \texttt{Loose Prompt-level Accuracy}: a prompt-level metric that permits minor formatting variations, such as extra whitespace or markdown artifacts.
\end{itemize}

Because the goal of this study is to evaluate agentic ensembles as deployable systems rather than only leaderboard entries, we additionally track operational and tokenomic characteristics:
\begin{itemize}
    \item \textbf{Cost per sample}: average API cost required to process one IFEval prompt.
    \item \textbf{Token composition}: the fraction of total tokens assigned to input context, accepted final output (goodput), exploratory sampled outputs, and orchestration/judging.
    \item \textbf{Goodput ratio}: the ratio of useful final-output tokens to total compute tokens consumed by the system.
    \item \textbf{Iteration progress}: cumulative accuracy after each feedback attempt, measuring the marginal utility of the judge-feedback loop.
    \item \textbf{Latency decomposition}: active processing time separated into model inference and orchestration/judging components.
    \item \textbf{Slice-level robustness}: accuracy by IFEval instruction category and by the number of constraints per prompt.
\end{itemize}
These measurements allow us to ask not only whether an ensemble improves accuracy, but also how much additional compute and orchestration overhead are required to obtain that improvement.

\section{Experimental Results}

We evaluate our ensembles on the IFEval benchmark. To estimate stochastic variation, each configuration is run five times and reported as mean accuracy with standard deviation ($\mu \pm \sigma$). Mistral and Llama variants were served via the OpenRouter platform~\cite{openrouter2026}. We first establish whether the proposed agentic SLM ensembles are competitive under standard IFEval metrics, and then analyze the tokenomic and operational costs of achieving these gains.

\begin{figure*}[p] % Changed [t] to [p] to give this massive stack a dedicated page
  \centering
  % --- FIRST FIGURE (Top) ---
  \includegraphics[width=0.91\linewidth, keepaspectratio]{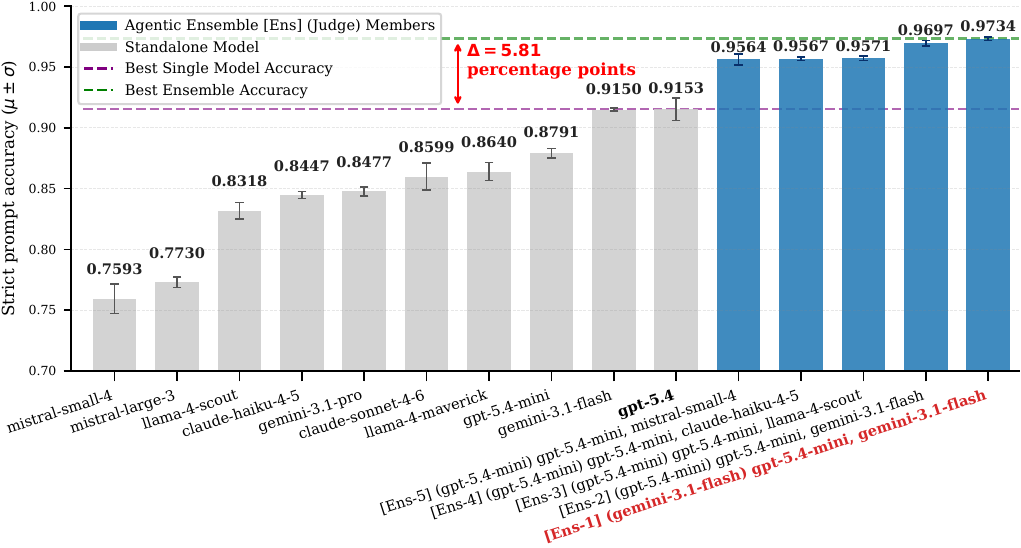}
  \caption{Strict prompt accuracy ($\mu \pm \sigma$) on the IFEval benchmark ($N=541$). For ensembles, identifiers denote the judge model in parentheses followed by the ensemble members. Horizontal dashed lines and bold labels denote the performance winners in each category, highlighting a gain of 5.81 percentage points for the proposed ensemble model over the strongest single-model baseline.}
  \label{fig:strict_accuracy}
  \Description{strict prompt accuracy}

  \vspace{1.5em} % Added a bit more breathing room between these two massive plots

  % --- SECOND FIGURE (Bottom) ---
  \includegraphics[width=0.91\linewidth, keepaspectratio]{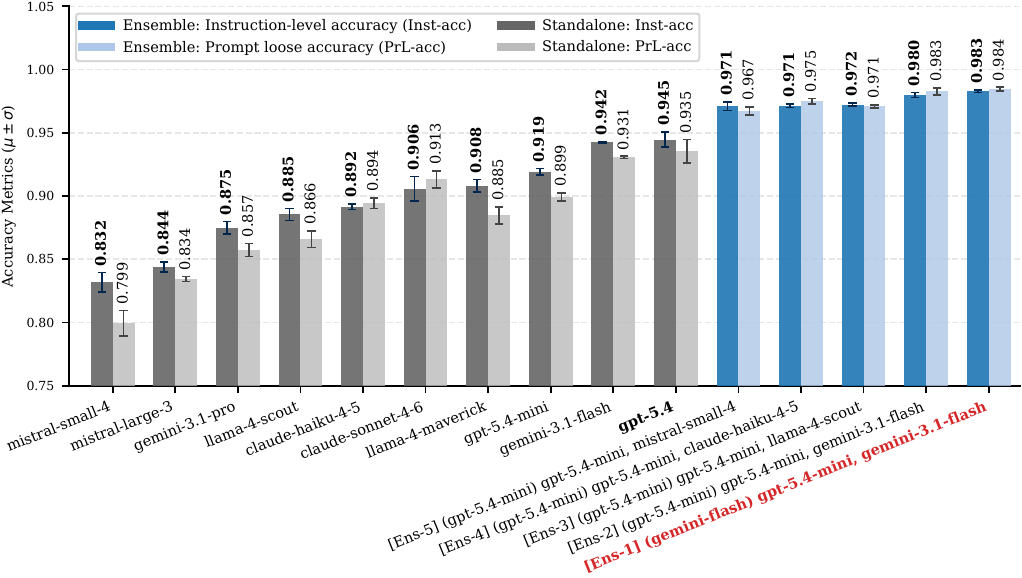}
  \caption{Instruction-level strict accuracy (Inst-acc) and prompt-level loose accuracy (PrL-acc) across standalone models and ensembles ($\mu \pm \sigma$) on the IFEval benchmark ($N=541$). Ensembles achieve higher overall accuracy and a significantly narrower gap between accuracy metrics, indicating ensemble robustness.}
  \label{fig:loose_instr_accuracy}
  \Description{prompt loose and instruction level accuracy}
\end{figure*}

Figure~\ref{fig:strict_accuracy} shows that all five ensemble configurations outperform the standalone baselines on strict prompt accuracy. The lowest-performing ensemble, [Ens-5], reaches $95.64\%$, while the strongest standalone model, gpt-5.4, reaches $91.53\%$. The best configuration, [Ens-1] with gemini-3.1-flash as judge and gpt-5.4-mini/gemini-3.1-flash as members, achieves $97.34\%$ strict prompt accuracy, $5.81$ percentage points above the strongest standalone baseline.

The same pattern holds for instruction and loose prompt metrics, where [Ens-1] reaches $98.3\%$ instruction-level accuracy and $98.4\%$ loose prompt-level accuracy. These complementary metrics, shown in Figure ~\ref{fig:loose_instr_accuracy}, indicate that the gains are not limited to a single strict formatting definition. Together, the standard IFEval metrics establish the viability of the agentic SLM ensemble architecture.

After demonstrating that the proposed SLM ensembles work under standard IFEval metrics, the central evaluation question becomes what operational tradeoffs enable these gains. A leaderboard score alone does not reveal whether the improvement comes from efficient collaboration, repeated sampling, judge overhead, or feedback-loop recovery. We therefore turn to beyond-leaderboard analysis, examining cost, token allocation, useful output goodput, iteration dynamics, latency, and robustness.

\begin{figure*}[p] % Changed [t] to [p] so this full 2x2 panel gets its own consecutive page
  \centering
  
  % --- Row 1, Left: Economic Frontier ---
  \begin{minipage}{0.48\textwidth}
    \centering
    \includegraphics[width=\linewidth]{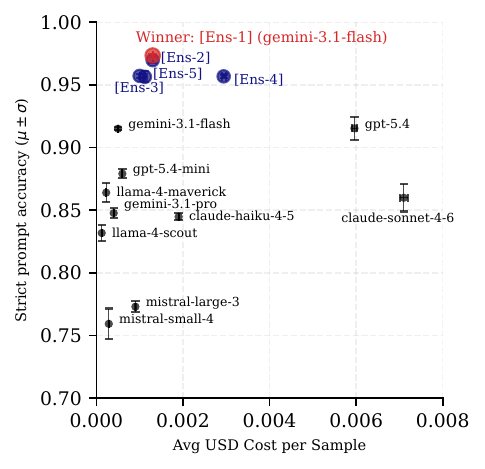}
    \captionof{figure}{Strict prompt accuracy vs. average cost per sample ($\mu \pm \sigma$). The winning ensemble [Ens-1] (red) with gemini-3.1-flash as a judge achieves the highest accuracy while maintaining a cost advantage over large models like gpt-5.4. The ensemble indices ([Ens-$n$]) match the ensembles in Figure 1.}
    \label{fig:economic_frontier}
  \end{minipage}
  \hfill
  % --- Row 1, Right: Token Composition ---
  \begin{minipage}{0.48\textwidth}
    \centering
    \includegraphics[width=\linewidth]{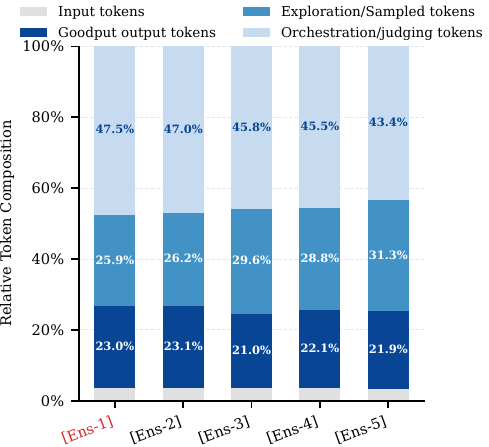}
    \captionof{figure}{Relative token composition per processing sample/prompt. The breakdown illustrates the average allocation of tokens across different ensembles. The ensemble indices ([Ens-$n$]) match the ensembles in Figure 1.}
    \label{fig:tax_breakdown}
  \end{minipage}

  \vspace{1.5em} % Balanced vertical spacing between the rows

  % --- Row 2, Left: Iteration Progress ---
  \begin{minipage}{0.48\textwidth}
    \centering
    \includegraphics[width=\linewidth]{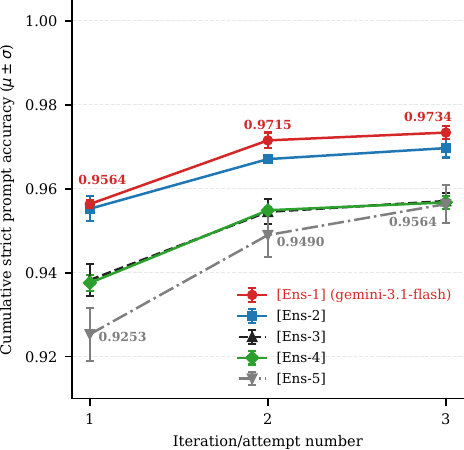}
    \captionof{figure}{The cumulative prompt-level strict accuracy progress over iterations. The upward trajectory demonstrates the system's capacity to identify instruction violations and provide corrective feedback to ensemble members, recovering valid outputs. The ensemble indices ($\text{[Ens-}n\text{]}$) match the ensembles in Figure 1.}
    \label{fig:iteration_progress}
  \end{minipage}
  \hfill
  % --- Row 2, Right: Intelligence Density ---
  \begin{minipage}{0.48\textwidth}
    \centering
    \includegraphics[width=\linewidth]{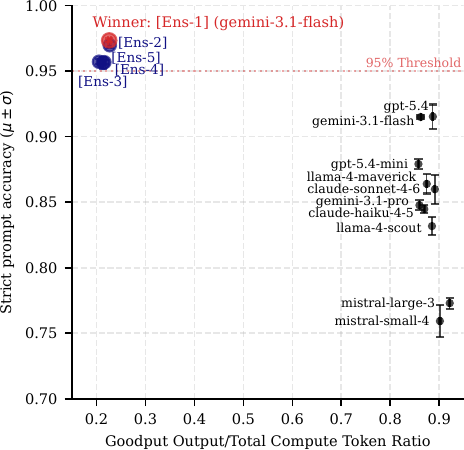}
    \captionof{figure}{Strict prompt accuracy vs. Goodput Output/Total Compute Token Ratio. The plot highlights ensembles clustering in the top-left quadrant, surpassing the 95\% accuracy threshold. This demonstrates that the ensemble of smaller models can achieve higher accuracy than a single large model, even though the ensemble's token overhead is higher. }
    \label{fig:intelligence_density}
  \end{minipage}
\end{figure*}

Figure~\ref{fig:economic_frontier} shows that the ensembles occupy a favorable region of the cost-performance space: they exceed the accuracy of the strongest standalone LLM while remaining less expensive per prompt than several large-model baselines. Figure~\ref{fig:tax_breakdown} shows token allocation breakdown between input, output, exploration, and orchestration overhead. The fact that orchestration tokens ($\sim$45–47$\%$) outweigh exploration tokens ($\sim$25–31$\%$) proves that the system is not just relying on sampling to find a "lucky" output. Instead, it spends more compute on analyzing and correcting errors than it is on sampling. This strongly supports the agentic self-correction narrative and shows that crossing the 95$\%$ accuracy threshold requires investment in orchestration, evaluation and structured feedback generation rather than just letting models blindly generate results in an open loop (zero--shot). Combined with Figure~\ref{fig:economic_frontier}, this shows that SLM ensembles can cost less than standalone LLMs despite orchestration overhead, owing to compute-efficient mini-model inference.

The feedback loop also contributes measurable recovery (Figure~\ref{fig:iteration_progress}). Across ensembles, cumulative strict accuracy increases over successive attempts, indicating that the judge-feedback mechanism contributes to the  recovery of initially invalid outputs. Finally, Figure~\ref{fig:intelligence_density} shows strict prompt accuracy plotted against the ratio of useful output tokens to total generated compute tokens, exposing two completely different clusters with no middle ground. Our ensembles cluster in the top-left quadrant, surpassing the $95\%$ accuracy threshold. Conversely, standalone models cluster heavily on the far right ($0.70\text{--}0.90$ token ratio) and hit a performance ceiling below the $92\%$ mark. This demonstrates that an ensemble of SLMs can outperform a single large model despite token overhead. 

\section{Conclusions and Future Work}
This work evaluates agentic SLM ensembles on IFEval from both a leaderboard and beyond-leaderboard perspective. 
The best ensemble achieves $97.34\%$ strict prompt accuracy, exceeding the strongest standalone LLM baseline by $5.81$ percentage points. This demonstrates that orchestrated SLMs, when coupled with programmatic IFEval checking and SLM-judge-mediated feedback, can outperform standalone LLMs on instruction-following tasks. More importantly, our tokenomic analysis shows that this improvement is enabled by tradeoffs: orchestration/judging overhead, iterative recovery, and latency costs. These results support the need for multi-dimensional evaluation of agentic systems, where accuracy, cost, token goodput, latency, and robustness are evaluated jointly. Future work will extend the framework to richer consensus mechanisms, including multi-agent panel debates and LLM-as-a-Jury methods, and to larger model pools. We also plan to study science-oriented agentic workflows, where reliable constraint following may extend beyond schema compliance to domain-specific validity checks.

\begin{acks}
Research presented in this report was supported by the Laboratory Directed Research and Development program of
Los Alamos National Laboratory under project numbers 20250637DI, 20250638DI, and 20250639DI.
\end{acks}

% ==========================================================
% PAGE 1: VERTICAL STACKED FIGURES
% ==========================================================

\clearpage

% ==========================================================
% PAGE 2: 2x2 PANELS
% ==========================================================

\clearpage % Flushes the 2x2 grid to Page 2 immediately before normal text resumes.
%  ======================================================================

\bibliographystyle{ACM-Reference-Format}
\bibliography{agentic-ensemble-mcp-ifeval-202605-v01}

\appendix

\section{Appendix}
\subsection{Model and Ensemble Configurations}
Model and ensemble configurations are shown in Table~\ref{tab:ensemble_configs}.

\begin{table*}[t]
  \caption{Evaluated large and small language models (LLMs and SLMs) and ensembles of SLMs.}
  \label{tab:ensemble_configs}
  \centering
  \begin{tabular}{l @{\hspace{0.25em}} ll}
    \toprule
    \textbf{Large Language Model (LLM)} & \multicolumn{2}{l}{\textbf{Small Language Model (SLM)}} \\
    \midrule
    gpt-5.4          & \multicolumn{2}{l}{gpt-5.4-mini} \\
    gemini-3.1-pro-preview   & \multicolumn{2}{l}{gemini-3.1-flash-lite-preview} \\
    claude-sonnet-4-6 & \multicolumn{2}{l}{claude-haiku-4-5} \\
    llama-4-maverick & \multicolumn{2}{l}{llama-4-scout} \\
    mistral-large-3  & \multicolumn{2}{l}{mistral-small-4} \\

    \midrule
    \multicolumn{3}{c}{\textbf{Ensemble composition: [Ens-n] (Judge model) ensemble member names}} \\
    \midrule
    {[Ens-1]} & (gemini-3.1-flash-lite-preview) & gpt-5.4-mini, gemini-3.1-flash-lite-preview \\
    {[Ens-2]} & (gpt-5.4-mini)     & gpt-5.4-mini, gemini-3.1-flash-lite-preview \\
    {[Ens-3]} & (gpt-5.4-mini)     & gpt-5.4-mini, llama-4-scout \\
    {[Ens-4]} & (gpt-5.4-mini)     & gpt-5.4-mini, claude-haiku-4-5 \\
    {[Ens-5]} & (gpt-5.4-mini)     & gpt-5.4-mini, mistral-small-4 \\
    \bottomrule
  \end{tabular}
\end{table*}

\subsection{Category-Level and Constraint-Complexity Analysis}

\begin{figure*}[p] 
  \centering
  \includegraphics[width=0.91\linewidth, keepaspectratio]{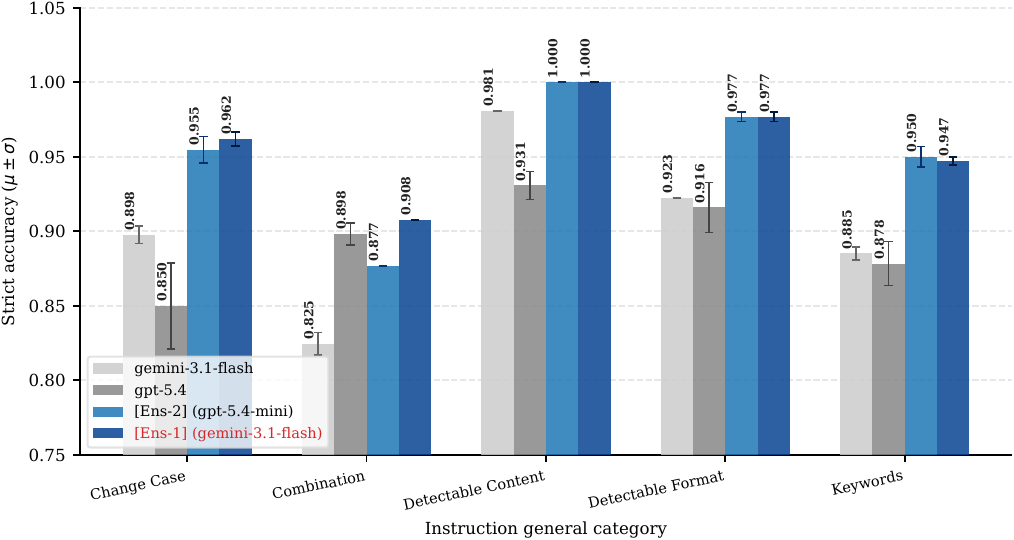}
  \label{fig:strict_acc_category_breakdown_a}
  \centerline{\small (a)}

  \vspace{1.5em} 

  \includegraphics[width=0.91\linewidth, keepaspectratio]{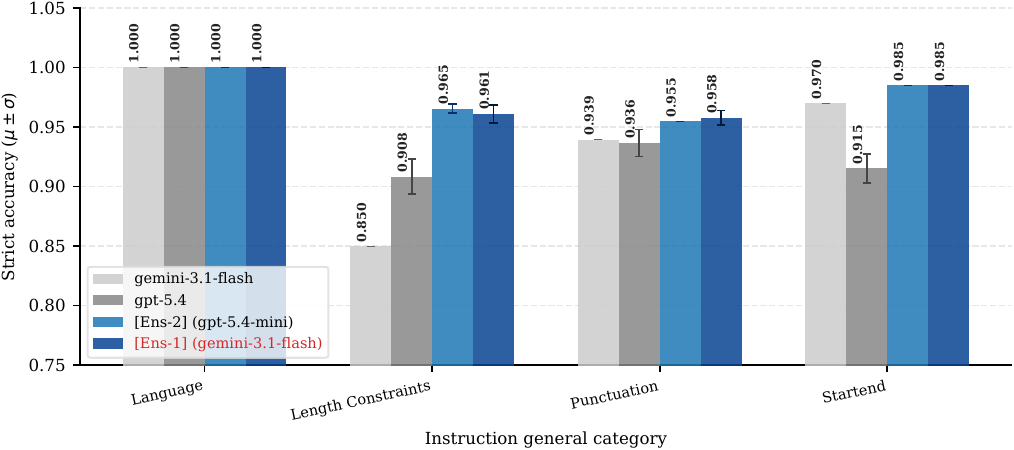}
  \centerline{\small (b)}

  \caption{Accuracy ($\mu \pm \sigma$) breakdown by IFEval instruction general categories. The breakdown is partitioned into (a) and (b). We show the top two standalone baselinesand two leading ensembles. The ensemble indices ($\text{[Ens-}n\text{]}$) match the ensembles in Figure 1.}
  \label{fig:strict_acc_category_breakdown}
  \Description{strict accuracy per category}
\end{figure*}

Figure~\ref{fig:strict_acc_category_breakdown} reports strict prompt accuracy across IFEval general instruction categories. The results show that ensemble gains are not uniform across instruction types: the largest benefits occur in categories where failures are often recoverable through feedback, such as formatting, content-detection, and multi-condition constraints. This supports the interpretation that the feedback loop improves instruction fidelity by correcting identifiable constraint violations rather than merely increasing generic model capability.

Figure~\ref{fig:accuracy_vs_constraints} further evaluates performance as a function of the number of instructions per prompt. As the number of simultaneous constraints increases, standalone model accuracy decreases more sharply, whereas the agentic ensembles retain higher strict accuracy. This suggests that ensemble feedback is particularly useful in compositional instruction-following settings, where a response may partially satisfy the prompt but fail one or more atomic constraints.

\subsection{Latency Decomposition}

\begin{figure*}[t] 
  \centering
  
  % --- Left Column: Constraint Complexity ---
  \begin{minipage}{0.48\textwidth}
    \centering
    \includegraphics[width=\linewidth]{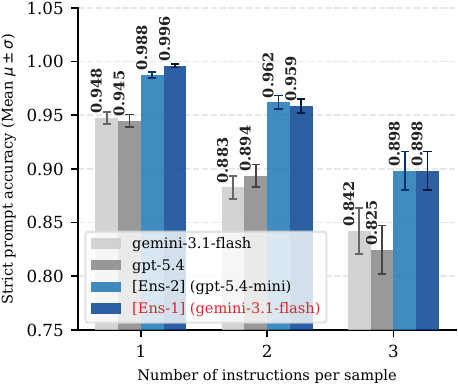}
    \caption{Strict prompt accuracy vs. sample complexity (defined as the number of instructions per prompt). We compare the top two standalone baselines with the leading ensembles. The results show that ensembles maintain higher accuracy at maximum complexity compared to standalone models. The ensemble indices ($\text{[Ens-}n\text{]}$) match the ensembles in Figure 1.}
    \label{fig:accuracy_vs_constraints}
    \Description{Strict prompt accuracy vs n instructions per sample.}
  \end{minipage}
  \hfill
  % --- Right Column: Latency Component Stack ---
  \begin{minipage}{0.48\textwidth}
    \centering
    \includegraphics[width=\linewidth]{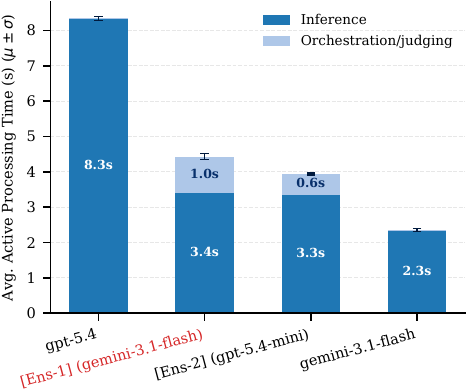}
    \caption{Average active processing time breakdown across the top two standalone models and the leading ensembles ($\mu \pm \sigma$) evaluated under sequential execution ($\text{max\_concurrency} = 1$). Total latency is partitioned into inference and orchestration/judging cycles.}
    \label{fig:active_latency_breakdown}
    \Description{Average active processing time stacked bar breakdown into inference and orchestration/judging segments for standalone and ensemble variants.}
  \end{minipage}
\end{figure*}

Figure~\ref{fig:active_latency_breakdown} decomposes average active processing time into inference and orchestration/judging components under sequential execution ($\text{max\_concurrency}=1$). This provides a conservative latency estimate because ensemble members and feedback cycles are not executed concurrently, although the implementation supports higher-concurrency runs. Even in this sequential setting, the leading SLM ensembles remain practical: they introduce judge and feedback overhead, but still achieve lower active processing time than the strongest standalone LLM baseline, gpt-5.4, while improving strict instruction-following accuracy. This highlights a key tokenomics tradeoff: additional coordination tokens and processing steps can be worthwhile when they are spent on compute-efficient component models.

\subsection{The Non-ASCII Roulette Anomaly}

During our systematic analysis of the underlying IFEval testing suite, we identified a small source of stochasticity in the benchmark's native evaluation routine, which we term the ``Non-ASCII Roulette'' anomaly. Specifically, the evaluation script applies a character replacement preprocessing step in which selected non-ASCII characters, as well as certain other characters whose code points fall outside predefined ranges, may be randomly replaced with standard ASCII characters. Importantly, this preprocessing is applied to the instruction constraints, not to the model output. For example, in a character-counting instruction such as ``the character `\#' should appear at least 3 times,'' the specified target character may be randomly replaced before compliance is checked.

As a result, the checker remains rule-based in its pass/fail logic, but the instruction preprocessing step can introduce randomness into what is being checked for affected character-based instructions. Consequently, repeated evaluations of the same model output may produce slightly different evaluation scores when the corresponding instruction contains affected characters.

To preserve comparability with existing IFEval leaderboards and prior reported results, we maintained the original evaluation parameters unchanged. However, this implementation detail should be considered when interpreting small differences in instruction- and prompt-level metrics, particularly for instructions involving affected non-ASCII or special characters.

% --- PLOT 1: Double-Column Category Breakdown (Takes Page 1 completely) ---

% --- PLOT 2: Side-by-Side Single-Column Frontiers (Queues for top of Page 2) ---

\end{document}